\documentclass[runningheads]{llncs}
\usepackage[T1]{fontenc}
\usepackage{graphicx}
\usepackage{comment}
\usepackage{makecell,multirow,diagbox}
\usepackage{amsmath,amssymb} 
\usepackage{color}
\usepackage{url}
\usepackage[misc]{ifsym}

\usepackage[breaklinks=true,bookmarks=false]{hyperref}

\begin{document}
\newcommand{\etal}{\textit{et al}.}
\newcommand{\ie}{\textit{i}.\textit{e}.}
\newcommand{\eg}{\textit{e}.\textit{g}.}
\pagestyle{headings}
\mainmatter
\def\ECCVSubNumber{1426}  

\title{CooGAN: A Memory-Efficient Framework for High-Resolution Facial Attribute Editing}

\titlerunning{CooGAN: A Memory-Efficient Framework for HR Facial Attribute Editing}
%
\author{Xuanhong Chen\inst{1,2}\protect\footnotemark[1] \and
Bingbing Ni\inst{1,2}\protect\footnotemark[4] \and
Naiyuan Liu\inst{1}\protect\footnotemark[5] \and
Ziang Liu\inst{1} \and
Yiliu Jiang\inst{1} \and
Loc Truong\inst{1} \and
Qi Tian\inst{3}
}
%
\authorrunning{Xuanhong Chen et al.}
%
\institute{Shanghai Jiao Tong University, China\\
\email{\{chen19910528,nibingbing,acemenethil,jiangyiliu\}@sjtu.edu.cn}, \email{\{liunaiyuan27,ttanloc\}@gmail.com}
\and
Huawei HiSilicon, China\\
\email{\{chenxuanhong,nibingbing\}@hisilicon.com}
\and
Huawei, China\\
\email{tian.qi1@huawei.com}}

\maketitle
\renewcommand{\thefootnote}{\fnsymbol{footnote}}
\footnotetext[1]{Work done during an internship at Huawei HiSilicon.}
\footnotetext[4]{Corresponding author: Bingbing Ni.}
\footnotetext[5]{Contributed to the work while he was a research assistant at Shanghai Jiao Tong University.}

\begin{abstract}
In contrast to great success of memory-consuming face editing methods at a low resolution, to manipulate high-resolution (HR) facial images, \ie, typically larger than $768^2$ pixels, with very limited memory is still challenging.
This is due to the reasons of 1) intractable huge demand of memory; 2) inefficient multi-scale features fusion.
To address these issues, we propose a NOVEL pixel translation framework called \emph{Cooperative GAN}(CooGAN) for HR facial image editing.
This framework features a local path for fine-grained local facial patch generation (\ie, patch-level HR, LOW memory) and a global path for global low-resolution (LR) facial structure monitoring (\ie, image-level LR, LOW memory), which largely reduce memory requirements.
Both paths work in a cooperative manner under a local-to-global consistency objective (\ie, for smooth stitching).
In addition, we propose a lighter selective transfer unit for more efficient multi-scale features fusion, yielding higher fidelity facial attributes manipulation.
Extensive experiments on  CelebA-HQ well demonstrate the memory efficiency as well as the high image generation quality of the proposed framework.
\end{abstract}

\begin{figure*}[t]
\begin{center}
\includegraphics[width=1\linewidth]{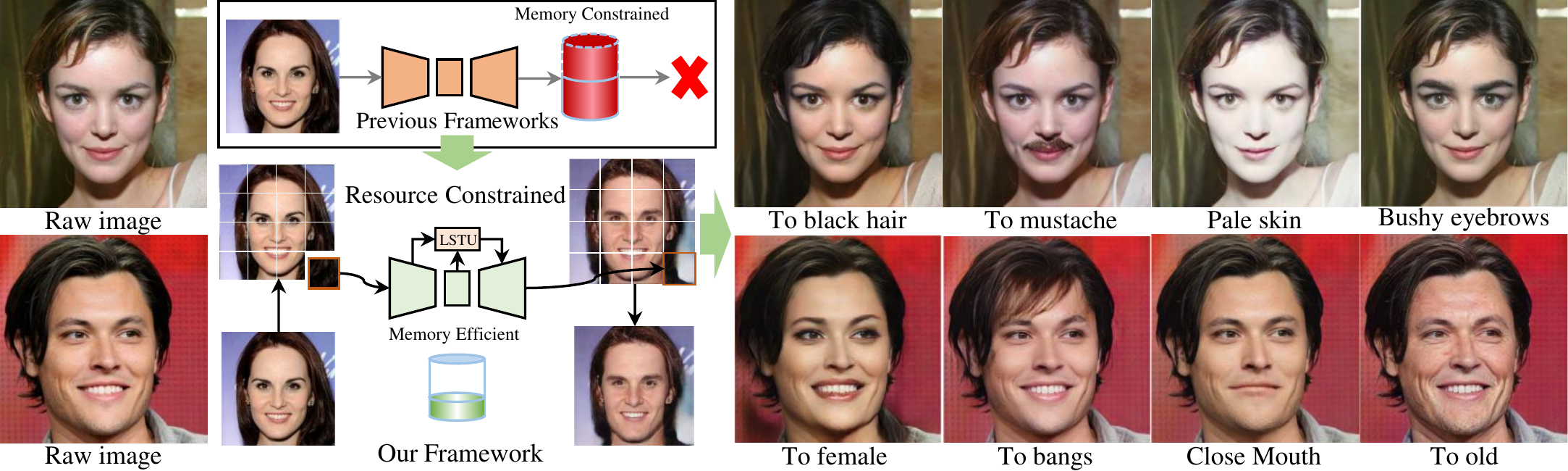}
\end{center}
\caption{For the purpose of translating the high-resolution images in the memory-constrained platform, we develop a memory-efficient image translation framework. In the figure, we show some high resolution ($768 \times 768$) editing results and its manipulation labels, more results can be found in the suppl. From the results, it can be seen that our framework can effectively edit both regional and global attributes.}
\label{motivation}
\end{figure*}

\section{Introduction}
Recent development in the field of deep learning has led to a rising interest in Facial Attribute Editing (FAE) ~\cite{DBLP:conf/cvpr/ChoiCKH0C18,DBLP:conf/aaai/Yin0L19,DBLP:journals/corr/abs-1908-07191,DBLP:journals/corr/abs-1907-11922,DBLP:journals/corr/abs-1905-01920,DBLP:journals/corr/abs-1908-08989,DBLP:journals/corr/abs-1904-06624,DBLP:conf/cvpr/0018DXLDZW19,DBLP:journals/corr/PerarnauWRA16}, \eg, manipulating structural, emotional, semantic attributes (\eg, gender, hair color, age etc.) of a given facial image.
These functionalities are highly demanded in emerging mobile applications.
However, most of recent facial editing approaches~\cite{DBLP:conf/cvpr/0018DXLDZW19,DBLP:journals/corr/abs-1907-11922,DBLP:journals/corr/abs-1908-07191} are based on deep image-to-image translation model, which can ONLY deal with low resolution (LR) facial images due to excessive deep model (\eg,  GAN~\cite{DBLP:conf/cvpr/XuZHZGH018,DBLP:conf/cvpr/0018DXLDZW19}) size.
Concretely, as the image size increases, the memory consumption of deep models also increases dramatically, even surpasses the limitation of device.
Such limitation forbids many promising applications on mobile platforms (\eg, ZAO~\cite{ZAO} processing images in the cloud service).
Although upgrading hardware can alleviate these issues, it is a non-cheap solution and still hard to process higher resolution images (\eg, 4K).
In this work, we aim at developing a memory-efficient deep framework to effectively handle high resolution facial attribute editing on resource-constrained devices.

There are two major challenges of deep model based HR facial attribute editing: 
1.~\emph{Constrained Computational and Memory Resource.}
In some mobile scenarios
(\eg, smartphone, AR/VR glasses) with only limited computational and memory resources, it is infeasible to use popular image editing models~\cite{DBLP:journals/corr/abs-1907-11922,DBLP:journals/corr/abs-1905-01920}  which require sophisticated networks.
To address this issue, methods based on model pruning and operator simplifying~\cite{DBLP:conf/cvpr/ZhaoNZZZT19,DBLP:journals/corr/HowardZCKWWAA17,DBLP:journals/corr/abs-1801-04381,DBLP:conf/cvpr/ZhangZLS18,DBLP:conf/eccv/MaZZS18} have been proposed to reduce  the inference computational complexity.
However, the metric-based way to reduce the model size will do harm to model perceptual representation ability, the output facial image quality is usually largely sacrificed.
2.~\emph{Inefficient Multi-scale Features Fusion.}
In order to achieve high-level semantic manipulation while maintaining local details during image generation, multi-scale features fusion is widely adopted.
It is a common practice to utilize skip connection, \eg,  U-Net~\cite{DBLP:conf/miccai/RonnebergerFB15}.
However, fixed schemes, such as skip connection, usually result in infeasible or even self-contradicting fusion output (\eg, during style transfer, content is well-preserved but failed to change the image style), and flexible schemes such as GRU~\cite{DBLP:journals/corr/ChungGCB14} lead to additional computational burden (\eg, applying GRU-based unit~\cite{DBLP:conf/cvpr/0018DXLDZW19} directly for multi-scale features fusion can achieve excellent fusion effects, but will increase network parameters by more than four times).

In this work, a novel image translation framework for high resolution facial attribute editing, dubbed \emph{CooGAN}, is proposed to explicitly address above issues.
It adopts a divide-and-combine strategy to break the \textbf{intractable} whole HR image generation task down to several sub-tasks for reducing memory cost greatly. 
More concretely, the pipeline of our framework consists of a series of local HR patch generation sub-tasks and a global LR image generation sub-task. To handle these two types of sub-tasks, our framework is also composed of two paths, \ie, local network path and global network path.
Namely, the local sub-task is to generate an HR patch with edited attributes and fine-grained details. And a global sub-task is to generate an LR whole facial snapshot with structural coordinates to guide the local workers properly recognize the correct patch semantics.
As only tiny size patch (\eg, $64\times 64$) generation sub-task is involved in the pipeline, the proposed framework avoids processing large size feature maps. As a result, this framework is very light-weighted and suited for resource constrained scenarios. 
In addition, a local-to-global consistency objective is proposed to enforce the cooperation of sub modules, which guarantees between-patch contextual consistency and appearance smoothness in fine-grained scale.

Moreover, we design a variant of SRU~\cite{DBLP:conf/emnlp/LeiZWDA18},~\emph{Light Selective Transfer Unit (LSTU)}, for multi-scale features fusion.
The GRU-based STU~\cite{DBLP:conf/cvpr/0018DXLDZW19} has similar functions, but needs two states (one state obtained from encoder, another from the higher level hidden state) to inference the selected skip feature.
As a result, it has to face a heavy computing burden and is not friendly to GPU acceleration.
Unlike STU, our SRU-based LSTU just need a single hidden state to get the gating signal, which greatly reduces the complexity of the unit.
Actually, the LSTU has only half as many parameters as STU and almost the same multi-scale features fusion effect, achieving a good balance between model efficiency and output image fidelity.
Under this design, the framework is able to selectively and efficiently transfer the shallow semantics from the encoder to decoder, enabling more effective multi-scale features fusion with constrained memory consumption.

We extensively experiment the proposed framework in terms of both qualitative and 
quantitative evaluations on facial attribute editing tasks.
It is demonstrated that our framework can process $768\times 768$ images well with less than $1GB$ GPU memory 
rather than over $6.3GB$ (the previous state-of-the-art~\cite{DBLP:conf/cvpr/0018DXLDZW19} level).
Furthermore, the proposed model has the capability to achieve \emph{$84.8\%$} facial attribute editing accuracy on CelebA-HQ.

\section{Related Works}
{\bf Image-to-image translation.} Image-to-image translation aims at learning cross-domain mapping in supervised or unsupervised settings~\cite{DBLP:conf/cvpr/0018DXLDZW19}. 
In the early stage,  methods like pix2pix use image-conditional GANs to train paired images in the supervised manner.
Later, unpaired image-to-image translation frameworks, which work in the unsupervised manner, are also proposed in~\cite{DBLP:conf/iccv/ZhuPIE17,DBLP:conf/nips/LiuBK17}. 
For instance, CycleGAN~\cite{DBLP:conf/iccv/ZhuPIE17} learns the mapping between one image distribution and another and introduces a cycle consistency loss to preserve key features during translation.
However, all those mentioned frameworks can only perform coarse-grain translation from one image distribution to another. They cannot modify specific attributes of an image, which is more important in real world applications. To address such problem, StarGAN~\cite{DBLP:conf/cvpr/ChoiCKH0C18} is proposed to learn multiple domain translation using only a single model.

Facial attribute editing is a typical task of image-to-image translation that focuses on editing specific attributes within image. Early methods~\cite{DBLP:conf/cvpr/Kemelmacher-ShlizermanSS14,DBLP:journals/tog/ThiesZNVST15,DBLP:conf/cvpr/ChenLSLTSYJ18} are proposed to learn only a single attribute editing model like expression transfer~\cite{DBLP:journals/tog/ThiesZNVST15}, age progression~\cite{DBLP:conf/cvpr/Kemelmacher-ShlizermanSS14}, and so on.
To achieve arbitrary attribute editing, IcGAN~\cite{DBLP:journals/corr/PerarnauWRA16} uses a conditional GAN to process encoded features. 
To strengthen the relations between high-feature space and attributes, Lample G. et al.~\cite{DBLP:conf/nips/LampleZUBDR17} applies adversarial constraints. 
StarGAN~\cite{DBLP:conf/cvpr/ChoiCKH0C18} and AttGAN~\cite{DBLP:journals/tip/HeZKSC19} use target attribute vector input and achieve great success.
Based on them, the STGAN~\cite{DBLP:conf/cvpr/0018DXLDZW19} is further proposed for simultaneously enhancing the attribute manipulation ability and image quality.
Especially, Ran Yi et al.~\cite{DBLP:conf/cvpr/YiLLR19} propose a hierarchical facial editing method to modify the face via facial features patches.
In our work, we analyze the limitation of StarGAN, AttGAN,~\cite{DBLP:conf/cvpr/YiLLR19} and STGAN and then further develop a patch-based model to make facial attribute editing suitable for higher resolution images.

\section{Methodology}
\subsection{Overview}
The proposed CooGAN framework for conditional facial generation presents two innovative modules, the global module and the local module. The global module is designed to generate LR translated facial image, and the local module aims at generating HR facial image patches and stitching them together. A cooperation mechanism is introduced to make these two modules work together, so that the global module provides the local module with a global-to-local spatial consistency constraint. 
In addition, to guarantee the performance and edit-ability of the generated images, we propose a well-designed unit, LSTU, to filter the features from latent space and infuse them with detail information inside the naive skip connection.

\subsection{The Cascaded Global-to-Local Face Translation Architecture}
\begin{figure}[t]
\centering
\includegraphics[width=1\linewidth]{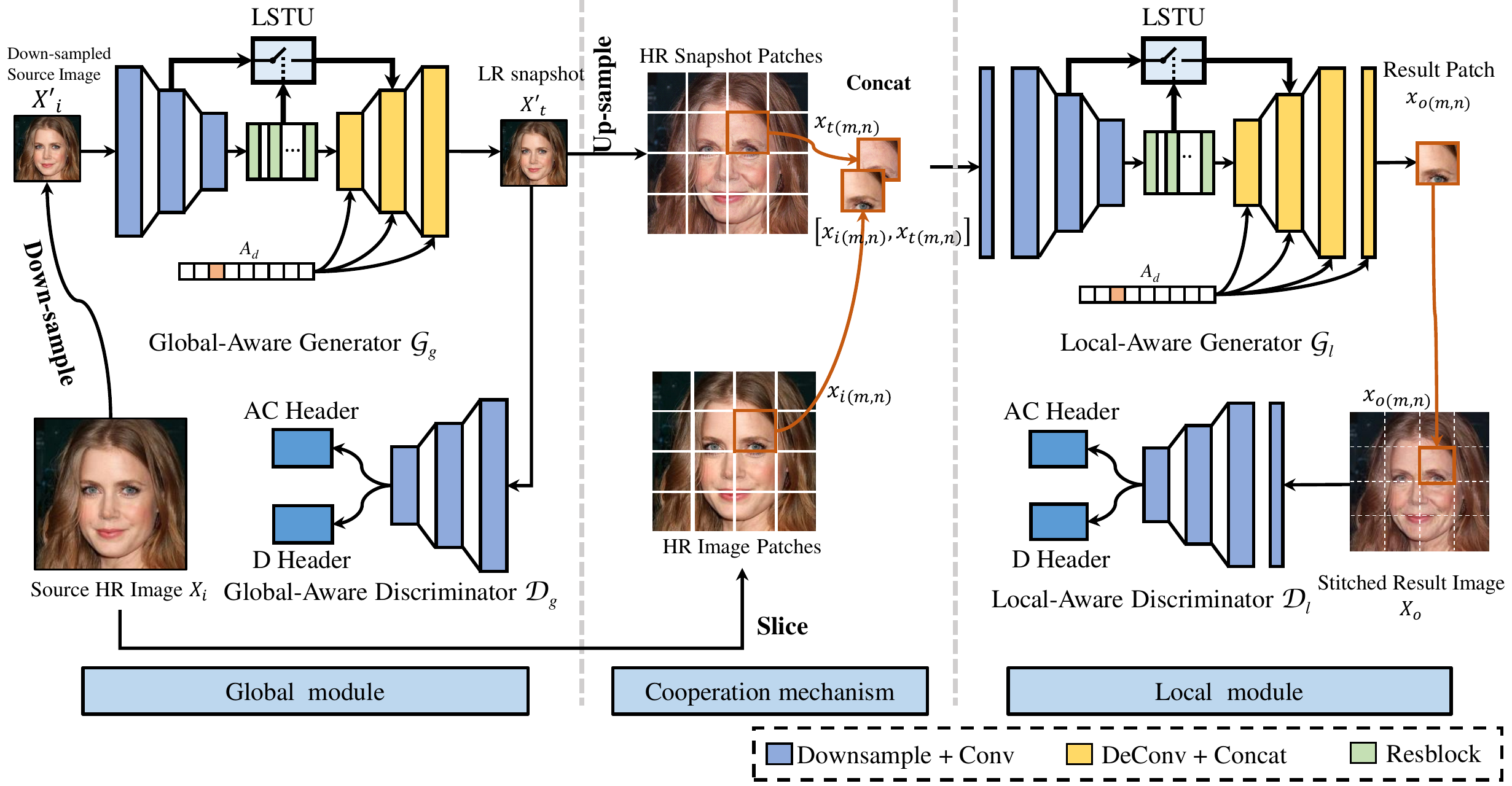}
\caption{An overview of our framework. Our framework has two modules--the global image translation module on the left side and the local patch refinement module on the right side. Either module contains one generator and one discriminator, as indicated in the graph. The two modules cooperate with each other through the cooperation mechanism in the middle. All specific operations are elaborated in the graph.}
\label{framework}
\end{figure}

The \emph{CooGAN} consists of two interdependent generation modules.
We depict the framework architecture in Fig.~\ref{framework}.

{\bf Global module.}
The global module is designed for generating the translated snapshot $X'_t$ which carries the whole image spatial coordinate information. 
As is shown in Fig.~\ref{framework}, this module contains two components: the global-aware generator $\mathcal{G}_g$ and the global-aware discriminator $\mathcal{D}_g$. 
$\mathcal{G}_g$ has a conventional U-Net~\cite{DBLP:conf/miccai/RonnebergerFB15} structure strengthening by LSTU. It takes in $X'_i$, the down-sampled image of $X_i$, and generates the LR snapshot $X'_t$.
Following the ACGAN-like~\cite{DBLP:conf/icml/OdenaOS17} fashion, the discriminator $\mathcal{D}_g$ has two headers sharing the same extracted features.
Note that the main purpose of global module is to guarantee the global semantic consistency of the final result. 

{\bf Local module.}
The local module is designed for processing HR image patches under the guidance of LR translated patches with limited computational resources.
As is shown in Fig.~\ref{framework}, the local module contains two components: the local-aware generator $\mathcal{G}_l$ and the local-aware discriminator $\mathcal{D}_l$.
We concatenate the original HR image patch and up-sampled snapshot patch, \ie, $[x_{i}(m,n), x_{t}(m,n)]$, where $(m,n)$ represents the coordinate of the patch in the whole image.
Then the local-aware generator $\mathcal{G}_l$ processes the concatenated data sequentially to generate translated HR patches. The generated patches can be stitched together \emph{without overlap} to synthesize the final output $X_o$.
To avoid the inconsistency between generated patches, the local-aware discriminator $\mathcal{D}_l$ is introduced to make the final stitched output smooth and seamless. 
Note that the discriminator is just introduced temporarily for training our model, so it will not add any memory cost during the inference stage.

{\bf Cooperation mechanism.} In the previous sections, we discuss the functions of the local module and the global module. 
The global module requires more computational resources while the size of image increases, which seriously limits its application for generating HR images.
The local module decomposes the full-size image into patches and translates the patches into HR ones, which is suitable for processing HR image without adding any memory cost.
But the local module alone cannot correctly perceive the patch semantics and is inapplicable for editing local attributes, as can be seen in Fig.~\ref{localalone}.
To this end, we introduce a valid cooperation mechanism to encourage the two generators to cooperate well for achieving satisfying generation performance.
As shown in the middle of Fig.~\ref{framework}, the global snapshot $X'_t$ is up-sampled to the same size as HR image $X_i$. Then, we decompose $X'_t$ and $X_i$ into a series of patches and concatenate the patches with the same coordinate.
The global patch carries the global spatial coordinate information while the HR image contains detailed textural information.
By this way, we combine the global spatial information and detailed textural information together.
Namely, the global spatial information is used to make the generated local patch smoother and globally consistent, and the detailed textural information is used to preserve the quality of the generation.

\subsection{Objective Function}
\label{Objective Function}
Our framework contains two sets of adversarial training in global and local module, respectively.
For the sake of clarity, we first describe the inference stage of either module.
In the global module, we get the snapshot $X'_t$ by the generator $\mathcal{G}_g$ and then feed $X'_t$ into the discriminator $\mathcal{D}_g$. 
Following ~\cite{DBLP:conf/cvpr/0018DXLDZW19}, we use $A_{d}$ to facilitate the adversarial training, where $A_{d}$ denotes the difference attribute vector between the target $A_{t}$ and the source $A_{i}$.
The training process can be defined as:
\begin{equation}
    X'_t = \mathcal{G}_g\left(X'_i,A_{d}\right).
\end{equation}

In the local module, $x_{i(i,j)}$ denotes a patch of HR image $X_i$ and $x_{t(i,j)}$ denotes a patch of up-sampled snapshot image $X_t$. We use $[ \cdot, \cdot]$ to represent the channel-wise patch concatenation,
\begin{equation}
    x_{o(i,j)} = \mathcal{G}_l\left(\left[x_{i(i,j)},x_{t(i,j)}\right],A_{d}\right).
\end{equation}

{\bf Image reconstruction loss.} We apply the reconstruction loss in our framework to improve image quality and avoid as many editing miscues as possible. This loss is achieved in a self-supervision manner, \ie, we feed the framework with zero condition input, which can be depicted as:
\begin{equation}
    \mathcal{L}_{rec} = ||x -\mathcal{G}(x,0)||_1.
\end{equation}

{\bf Adversarial loss.} For FAE task, there is no ground-truth for training.
Adversarial loss is usually applied due to its domain-to-domain translation nature.
To alleviate the instability of adversarial loss in training process, gradient penalty~\cite{DBLP:journals/corr/GulrajaniAADC17} is introduced:

\begin{equation}
    \begin{split}
    \begin{aligned}
      \max \limits_{\mathcal{D}_{adv}} \mathcal{L}_{\mathcal{D}_{adv}} = \mathbb{E}_x\big[\mathcal{D}_{adv}(x)\big] - \mathbb{E}_{\hat{y} }\big[\mathcal{D}_{adv}({\hat{y}})\big] &+\qquad \\
        \lambda \mathbb{E}_{\hat{x}}\big[(&\|\nabla_{\hat{x}}\mathcal{D}_{adv}({\hat{x}})\|_2-1)^2\big],
    \end{aligned}
    \end{split}
\end{equation} 
\begin{equation}
    \max \limits_{\mathcal{G}} \mathcal{L}_{\mathcal{G}_{adv}} = \mathbb{E}_{x,A_{d}}\big[\mathcal{D}_{adv}(\mathcal{G}(x,A_{d}))\big],\quad\quad\quad\qquad\quad\quad\qquad\qquad 
\end{equation} 
where $\mathcal{D}_{adv}$ denotes the adversarial header of the discriminator. $x$ is the input real image, $\hat{y}$ is the generated image and $\hat{x}$ is the sampled point along straight lines between the real image distribution and generated image distribution.

{\bf Attribute editing loss.} As discussed above, attribute editing has no ground truth. Following the ideology of ACGAN~\cite{DBLP:conf/icml/OdenaOS17}, we add a classification header sharing features with adversarial header. Such a design effectively realizes adversarial training along with attribute learning. The attribute editing loss below supervises the two modules of our framework individually,
\begin{equation}
    \begin{split}
        \mathcal{L}_{\mathcal{D}_{a}} = 
        -\sum_{i=1}^{N_c}
        \left[a_t^{(i)}\log{\mathcal{D}_{a}^{(i)}(x)}+(1-a_t^{(i)})\log{(1-\mathcal{D}_{a}^{(i)}(x))} \right],
    \end{split}
\end{equation}
where $a^{(i)}_t$ represents the $i$-th target attribute and $\mathcal{D}_a$ represents the classification header.
We use the combination of the above three types of loss to train our model.

\begin{table}[b]
\tiny
\setlength{\tabcolsep}{5mm}
\begin{center}
\footnotesize
\label{parameters}
\begin{tabular}{cccccc}
\hline
SC type&SC$0$&SC$1$&SC$2$&SC$3$&SC$4$\\
\hline
PSNR&$20.2$&$24.29$&$28.0$&$29.9$&$34.8$\\
SSIM&$0.643$&$0.777$&$0.877$&$0.916$&$0.989$\\
Average Accuracy &$89.5\%$&$82.4\%$&$75.1\%$&$69.1\%$&$63.9\%$\\
\hline
\end{tabular}
\caption{PSNR/SSIM performance and average attribute editing accuracy of models with different Skip Connection (SC) numbers, SC$i$ represents model with $i$ skip connections.}
\label{PSNR}
\end{center}
\end{table}

\subsection{Light Selective Transfer Unit}
As mentioned above, we proposed a high-resolution facial image editing framework which can process HR images with limited computational resources.
In order to further improve the generation performance, we improve the manner of multi-scale features fusion.

The most popular method for multi-scale features fusion in image-to-image translation is the \emph{Skip Connection} structure. It helps the network balance the contradiction between the pursuit of larger receptive field and loss of more details. 
One of its classic deployments is U-Net.
However, there is a fatal drawback of the original skip connection: it will degrade the function of deeper parts and further damage the effectiveness of condition injection.
From Table~\ref{PSNR}, it is obvious that PSNR increases but attribute editing accuracy decreases when the skip connection number multiplies. A detailed graph showing the editing accuracy of each specific attribute is given in suppl.
STGAN~\cite{DBLP:conf/cvpr/0018DXLDZW19} tries to alleviate the problem with the STU, a variant of GRU~\cite{DBLP:conf/emnlp/ChoMGBBSB14,DBLP:journals/corr/ChungGCB14}, which uses the latent feature to control the information transfer in the skip connection through the unit.
This feature carries the conditional information added through the concatenation.
Unfortunately, such a unit omits the spatial and temporal complexity and it is a time-consuming process for the underlying feature to bubble up from the bottleneck to drive the STU of each layer.

\begin{figure}[t]
\begin{center}
\includegraphics[width=0.9\linewidth]{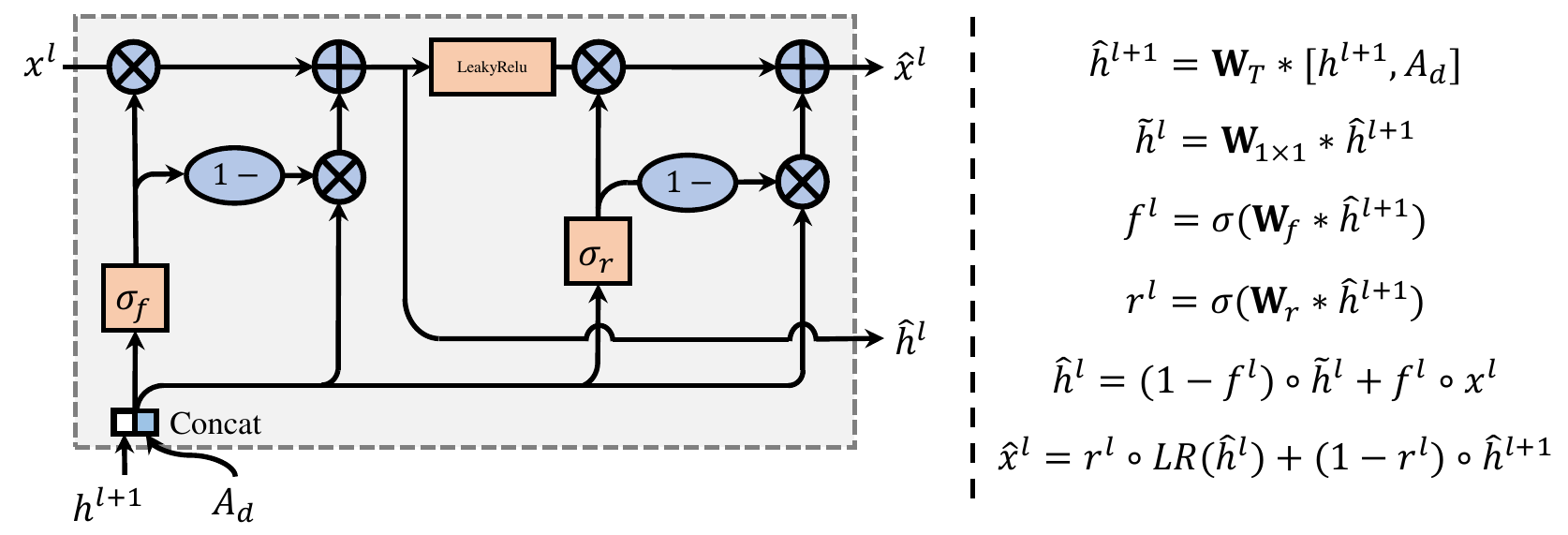}
\end{center}
  \caption{The structure of proposed LSTU. The design of LSTU is inspired by SRU, which makes LSTU more lightweight than STU and suitable for GPU parallel acceleration. The right side shows the mathematical expression of the inference process of our LSTU. LR is short for LeakyRelu.}
  \label{lstu}
\end{figure}

To explicitly address the mentioned problem, we present our framework to employ \emph{Light Selective Transfer Unit (LSTU)} to efficiently and selectively transfer the encoder feature.
LSTU is an SRU-based unit with totally different information flow.
Compare to STU, our LSTU discards the dependence on the two states when calculating the gating signal, which greatly reduces our parameters but the unit is still efficient.
The detailed structure of LSTU is shown in Fig.~\ref{lstu}. 
Without loss of generality, we choose the $l$-th LSTU as an analysis example. 
The $l$-th layer feature coming from the encoder side denotes as $x^l$.
$h^{l+1}$ denotes the feature in the adjacent deeper layer.
It contains the filtered latent state information of that layer .
$h^{l+1}$ is firstly concatenated with attribute difference $A_d$ to obtain up-sampled hidden state $\hat{h}^{l+1}$.
Then $\hat{h}^{l+1}$ is used to independently calculate the masks $f^l,r^l$ for the forget-gate and reset-gate.
$\mathbf{W_T}$, $\mathbf{W_{1\times 1}}$, $\mathbf{W_f}$ and $\mathbf{W_r}$ represent parameter matrix of transpose convolution, linear transform, forget gate and update gate.
The further process is similar to SRU. 
The equation of gates is shown on the right side of Fig.~\ref{lstu}.

\section{Experiments}
In this section, we take the following steps to evaluate the performance of proposed CooGAN:
firstly, we verify the effectiveness of the LSTU in multi-scale features fusion.
Then, we discuss the facial attribute editing performance and the memory usage of our overall framework.
\subsection{Implement Details}
{\bf Dataset.} We train our model on the CelebA~\cite{liu2015faceattributes} dataset (for the global module) and CelebA-HQ dataset (for the local module). CelebA is a collection of more than 200,000 human facial images and labels of 40 attributes for each of the images. CelebA-HQ is an artificial high resolution dataset translated from CelebA utilizing the method from~\cite{DBLP:conf/iclr/KarrasALL18}. 
Considering the distinguishability of attributes, in the experiments we mainly use 13 attributes including \emph{Bald, Bangs, Black Hair, Blond Hair, Brown Hair, Bushy Eyebrows, Eyeglasses, Male, Mouth Slightly Open, Mustache, No Beard, Pale Skin and Young}.

{\bf Optimizer.} The optimizer used in our training process is Adam~\cite{DBLP:journals/corr/KingmaB14} with hyper-parameter $\beta=(0.5,0.99)$ and learning rate $0.0002$.
And we apply the learning rate decay strategy with the decay rate 0.1 and decay epoch 100.

\begin{table}[b]
\begin{center}
\footnotesize
\setlength{\tabcolsep}{4.05mm}
\begin{tabular}{ccccc}
\hline
Method&StarGAN&AttGAN&STGAN&Ours\\
\hline
PSNR/SSIM&$22.86/0.848$&$24.1/0.821$&$31.6/0.934$&${\bf 32.1/0.938}$\\
\hline
\end{tabular}
\caption{Reconstruction PSNR and SSIM (CelebA $128\times 128$) of different facial attribute editing frameworks.}
\label{diffframeworkpsnr}
\footnotesize
\end{center}
\end{table}

{\bf Quantitative metric.} We design and train an attribute-classifier to quantify attribute editing effects in form of attribute classes accuracy.
It takes form of the classifying part in our discriminator and achieves a $96.0\%$ average accuracy on CelebA.
And for the texture sharpness of reconstruction results, Peak Signal to Noise Ratio(PSNR) and Structural Similarity(SSIM) are used. They are two common metrics to measure the generated image fidelity.
In addition, we use the floating-point operations per second (FLOPS) and model parameter size to evaluate the memory efficiency of networks.

\subsection{Experimental analysis of Framework.}
{\bf Experiment setup.}
Our framework consists of two modules: global module and local module.
In the following experiments, the input image size of the global module is set as $256 \times 256$
and the input patch size of local module is $128\times 128$. The final high resolution image size is $768\times 768$.
We use the two most effective previous methods (AttGAN, STGAN) and our global module as comparison objects.
Since these models are not originally designed for HR image processing, we make proper adjustments of increasing their layers (AttGAN-HR: 7 layers, STGAN-HR: 7 layers, LSTU-net-HR: 7 layers) to expand their perceptual fields. We expand the models based on their official codes and train them with their original loss functions. Those adjusted approaches can be denoted as: AttGAN-HR (AttGAN), STGAN-HR (STGAN) and LSTU-net-HR(global module).
The complete model structure setting of our framework, AttGAN-HR and STGAN-HR can be found in suppl.
Models are evaluated in the CelebA-HQ dataset.

\begin{figure}[t]
\begin{center}
\includegraphics[width=0.6\linewidth]{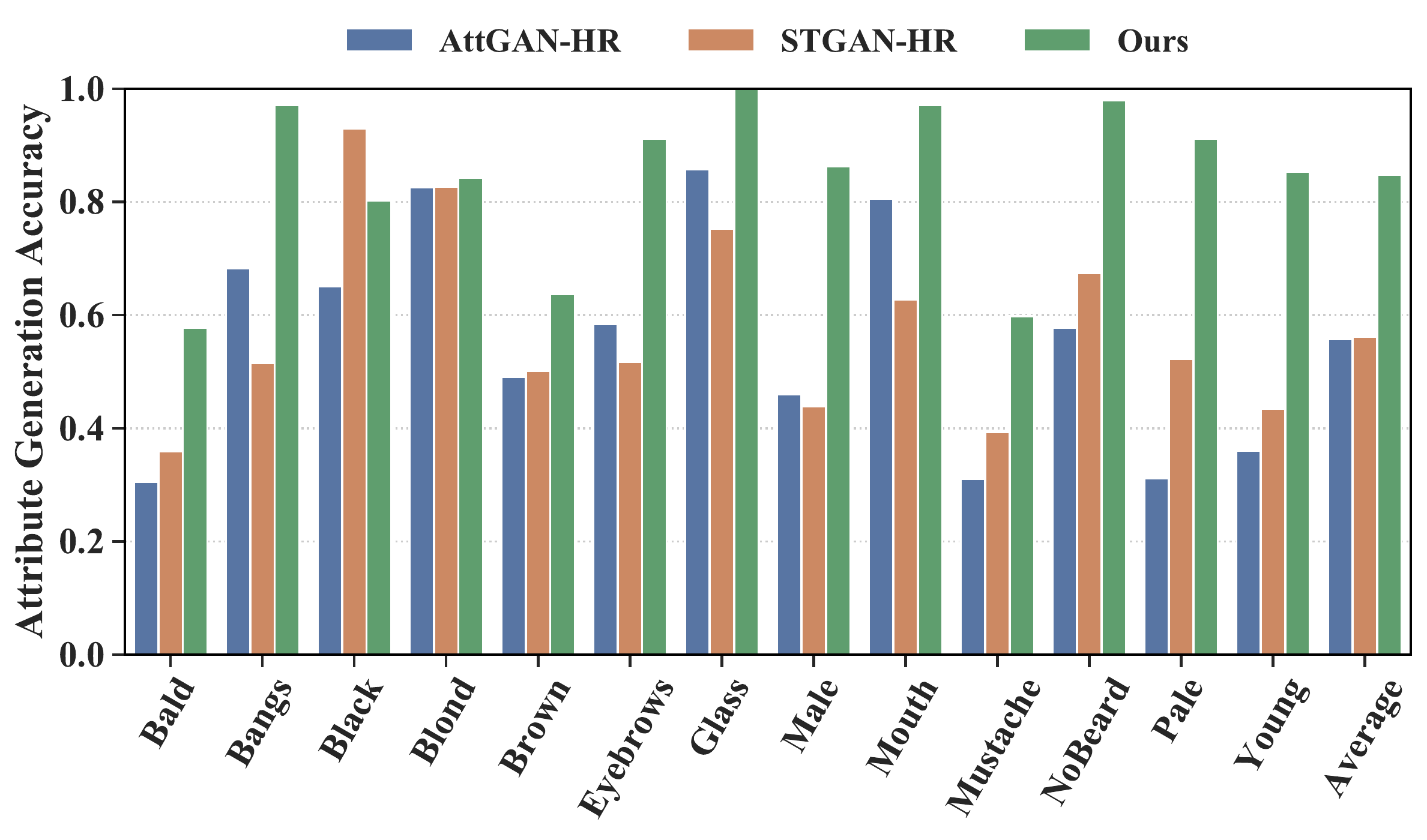}
\end{center}
  \caption{Our CooGAN against other approaches in attribute translation accuracy. Our framework transcend others in most attributes.}
\label{acc1_hr}
\end{figure}

\begin{table}[b]
\begin{center}
\setlength{\tabcolsep}{6.6mm}
\begin{tabular}{ccccc}
\hline
Method&AttGAN-HR&STGAN-HR&Ours\\
\hline
GPU Memory&$3692 $MB&$6306 $MB&${\bf 985} $MB\\
\hline
Inference time&$0.0414 $s&$0.0502 $s&${\bf 0.0645} $s\\
\hline
\end{tabular}
\caption{GPU memory consumption and average inference time (200 images), measured on a 2080ti, for different framework with batch size 1, input images at the $768\times 768$ pixels.}
\label{gpuconsumption}
\end{center}
\end{table}

\begin{figure}[!t]
\begin{center}
\includegraphics[width=1\linewidth]{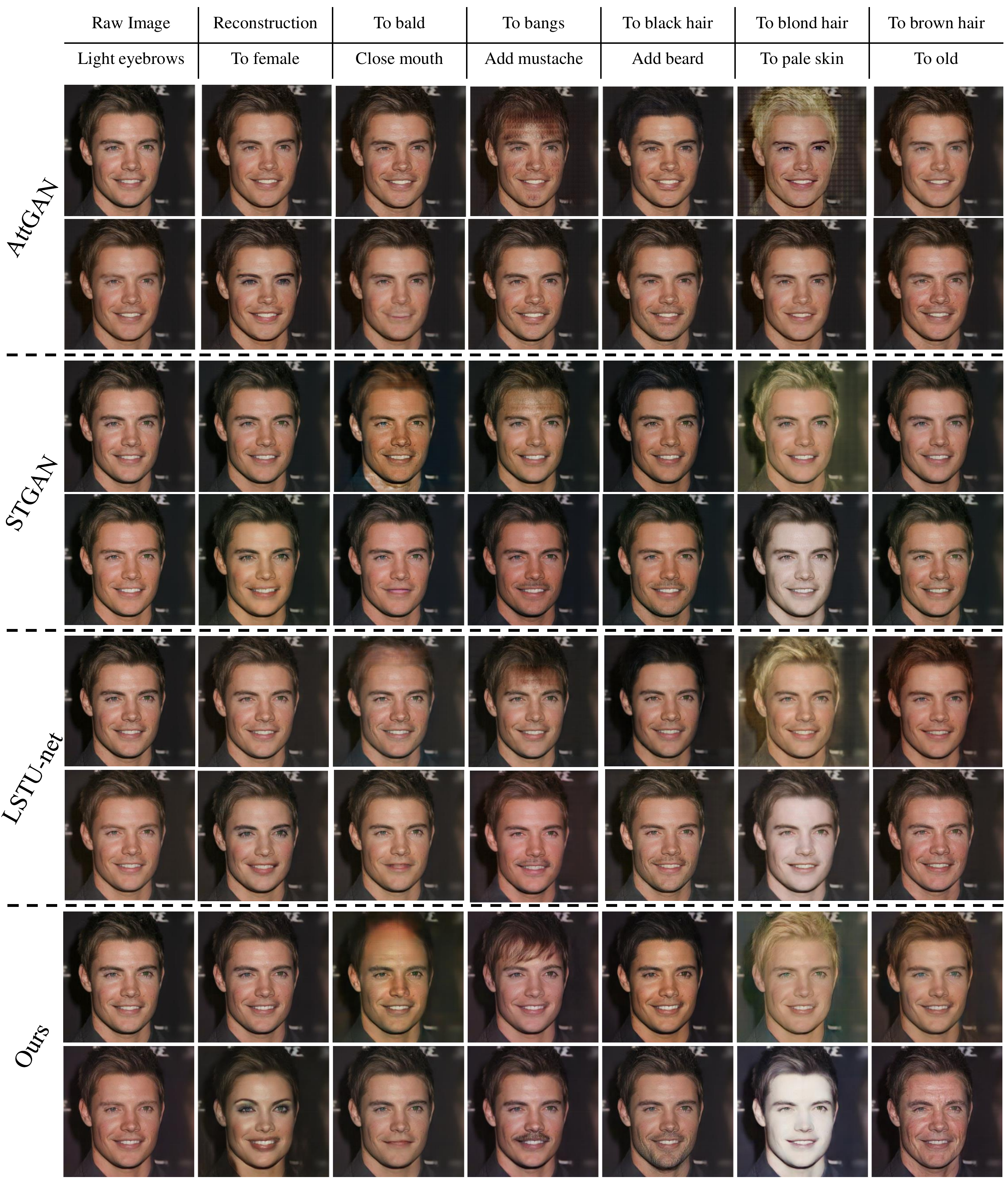}
\end{center}
\caption{High resolution ($768\times 768$) facial attribute editing results of AttGAN-HR, STGAN-HR, LSTU-net-HR and our framework. The textural details can be observed clearly through zooming in. More HR editing results can be found in suppl.}
\label{results}
\end{figure}

{\bf Quantitative evaluation.} We report the attribute editing accuracy of our framework and other methods in Fig.~\ref{acc1_hr}. Our framework can achieve $84.8\%$ in classification accuracy, 31.3 in PSNR and 0.923 in SSIM.
In terms of memory usage, the largest memory consumption occurs in the global module for the reason that the local module only processes image patches with considerably small sizes. As is shown in Table~\ref{gpuconsumption}, while processing $768\times 768$ images the maximum consumption of our framework is below all compared models, even StarGAN. This is on the grounds that the peak consumption in our framework happens when the global module processes $256\times 256$ images, while other models process $768\times 768$ images directly. Besides, our framework reduces large memory consumption at the cost of only a little inference time increase, as there is \emph{no overlap} when our framework processes patches, i.e., our advantage. Therefore, the prominent image processing efficiency of our framework is verified.

{\bf Qualitative analysis.} 
The visual effect comparison are shown in Fig.~\ref{results}.
In general, our CooGAN can effectively generate high-resolution images with smooth patch boundaries.  As is shown, images generated by our CooGAN possess more prominent characteristics as well as better textural quality. For local attributes like bald and mustache, our framework performs much more accurate and significant modification. And in global attributes like gender and age, our CooGAN translates the image more thoroughly and successfully. It is worth mentioning that HR translated images have better effect in some attributes like age and gender. This is because these attribute can not be properly manifested in LR images, \eg, the wrinkles of aging are obvious only in high resolution.

\begin{figure}[t]
\begin{center}
\includegraphics[width=0.8\linewidth]{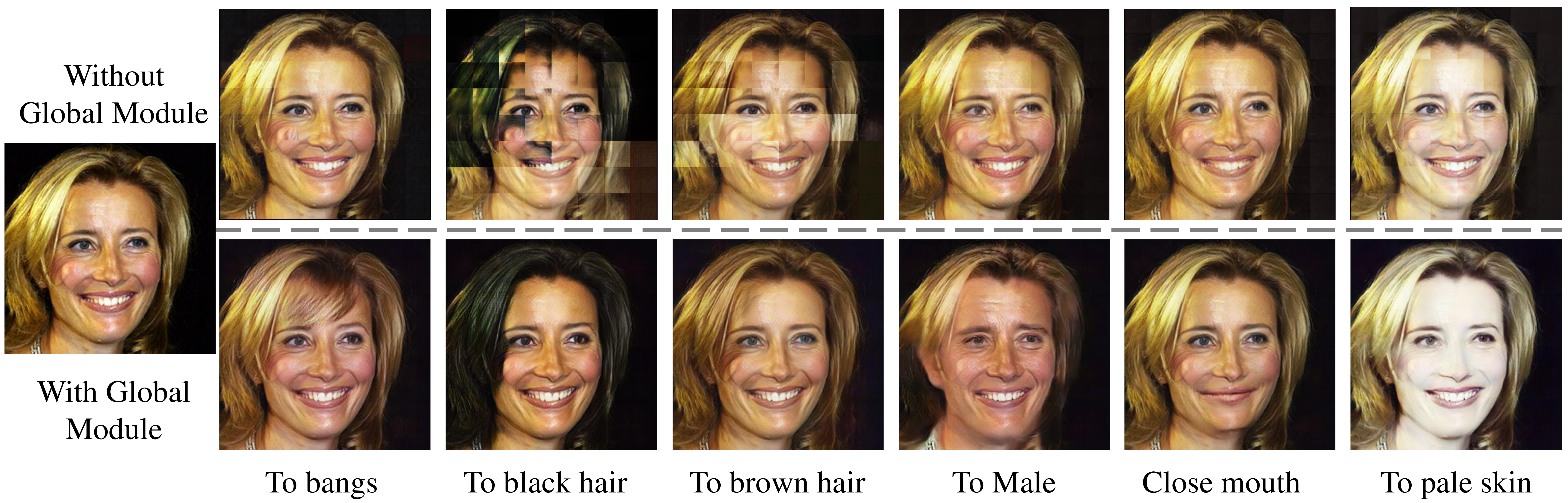}
\end{center}
\caption{$768\times 768$ results generated by our framework without and with global module.}
\label{localalone}
\end{figure}
\begin{figure}[t]
\begin{center}
\includegraphics[width=1\linewidth]{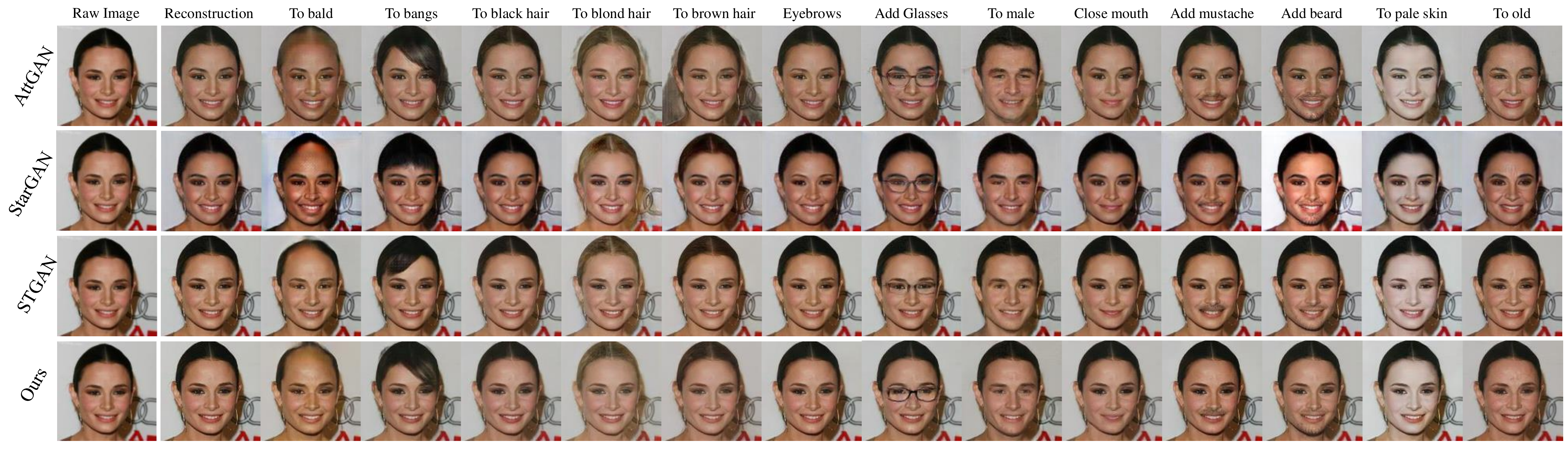}
\end{center}
\caption{Low resolution ($128\times 128$) facial attribute editing results of StarGAN, AttGAN, STGAN and our LSTU-Net. }
\label{lrresults}
\end{figure}

\begin{table}[b]
\scriptsize
\begin{center}
\setlength{\tabcolsep}{2.5mm}
\begin{tabular}{cccccccc}
\hline
Method&\emph{Bald}&\emph{Bangs}&\emph{Black}&\emph{Blond}&\emph{Brown}&\emph{Eyebrows}&\emph{Glasses}\\
\hline
AttGAN-HR& $25.7\%$&   $66.2\%$&   $61.2\% $&   $89.4\%$&   $47.9\%$&   $51.7\%$&   $86.7\%$\\
STGAN-HR&  $38.8\%$&   $50.5\%$&   $78.2\%$&   $92.8\%$&   $53.5\%$&   $65.4\%$&   $71.5\%$\\
Ours&$\bf61.5\%$&$\bf93.3\%$&$\bf90.6\%$&$\bf94.9\%$&$\bf73.6\%$&$\bf84.9\%$&$\bf94.8\%$\\
\hline
\end{tabular}
\setlength{\tabcolsep}{2.3mm}
\begin{tabular}{cccccccc}
\hline
Method&\emph{Male}&\emph{Mouth}&\emph{Mustache}&\emph{No Beard}&\emph{Pale}&\emph{Young}&\emph{Smooth}\\
\hline
AttGAN-HR&     $36.8\%$&   $ 83.2\%$&   $29.8\%$&   $ 51.4\%$&   $34.5\%$&   $41.4\%$&   $99.8\%$\\
STGAN-HR&$51.3\%$&   $67.7\%$&   $49.1\%$&   $67.8\%$&   $61.4\%$&   $49.9\%$&   $99.9\%$\\
Ours&$\bf87.9\%$&$\bf91.7\%$&$\bf61.9\%$&$\bf93.8\%$&$\bf98.1\%$&$\bf64.7\%$&$\bf91.5\%$\\
\hline
\end{tabular} 
\caption{User study results of our framework, STGAN-HR and AttGAN-HR.}
\label{userstudy}
\end{center}
\end{table}

{\bf User study.} To further validate the subjective performance of our framework, it is important to conduct survey on a crowdsourcing platform to acquire people's general opinions of our HR edited images and compare our CooGAN with the modified HR versions of previous state-of-the-art models including STGAN-HR and AttGAN-HR. In the survey, 50 questions of 14 attributes/quality are involved, \ie, 13 previous mentioned attributes and the smoothness of the image. And in each question, a randomly selected image and its edited versions by the three models are presented in random sequences. And people will determine whether an edited image has correct and realistic attributes and is generally smooth and consistent. The user study results are given in Table~\ref{userstudy}.
In comparison, our CooGAN shows its excellent performance in attributes editing. 
Because our framework generates image by patches, it is slightly inferior to the other two methods in terms of smoothness.

{\bf Ablation study for Cooperative mechanism.} In this part, our focus is to study the importance and effectiveness of global module in the framework. We train our framework with and without the global module. The results show that without global module the attribute editing will be impractical, as is shown in Fig.~\ref{localalone}. On the one hand, without the spatial information given by global snapshot, the local module can not well recognize local attributes like bangs regions and mouth. On the other hand, the shrinkage of perceptive field causes loss of global semantics, rendering certain features (like gender characteristics), difficult for the framework to discern. Thus, we confirm the irreplaceable role of global module in our framework.

\subsection{Ablation study for LSTU}
LSTU is a vital component of our framework.
We will verify the feasibility and effectiveness of LSTU in this section.

{\bf Experiment setup.} 
Our LSTU is a lightweight and equally effective replacement unit for STU~\cite{DBLP:conf/cvpr/0018DXLDZW19} (previous state-of-the-art multi-scale features fusion unit proposed in STGAN). 
To validate its effect and efficiency, we replicate the network structure of STGAN and replace its STU with our LSTU. We name this new network LSTU-Net to be distinguished from the original STGAN.
Besides, we also compare LSTU-Net with StarGAN and AttGAN since they are the most relative and effective facial attribute editing methods that can be found.
However, these compared models are not designed for HR image generation tasks. They are mostly purposed to process $128\times 128$ images.
To be fair, all models are tested on $128\times 128$ CelebA dataset.

In addition, the official codes and pretrained models of those three compared models are used in our experiments for the sake of impartiality.

\begin{figure}[t]
\begin{center}
\includegraphics[width=0.6\linewidth]{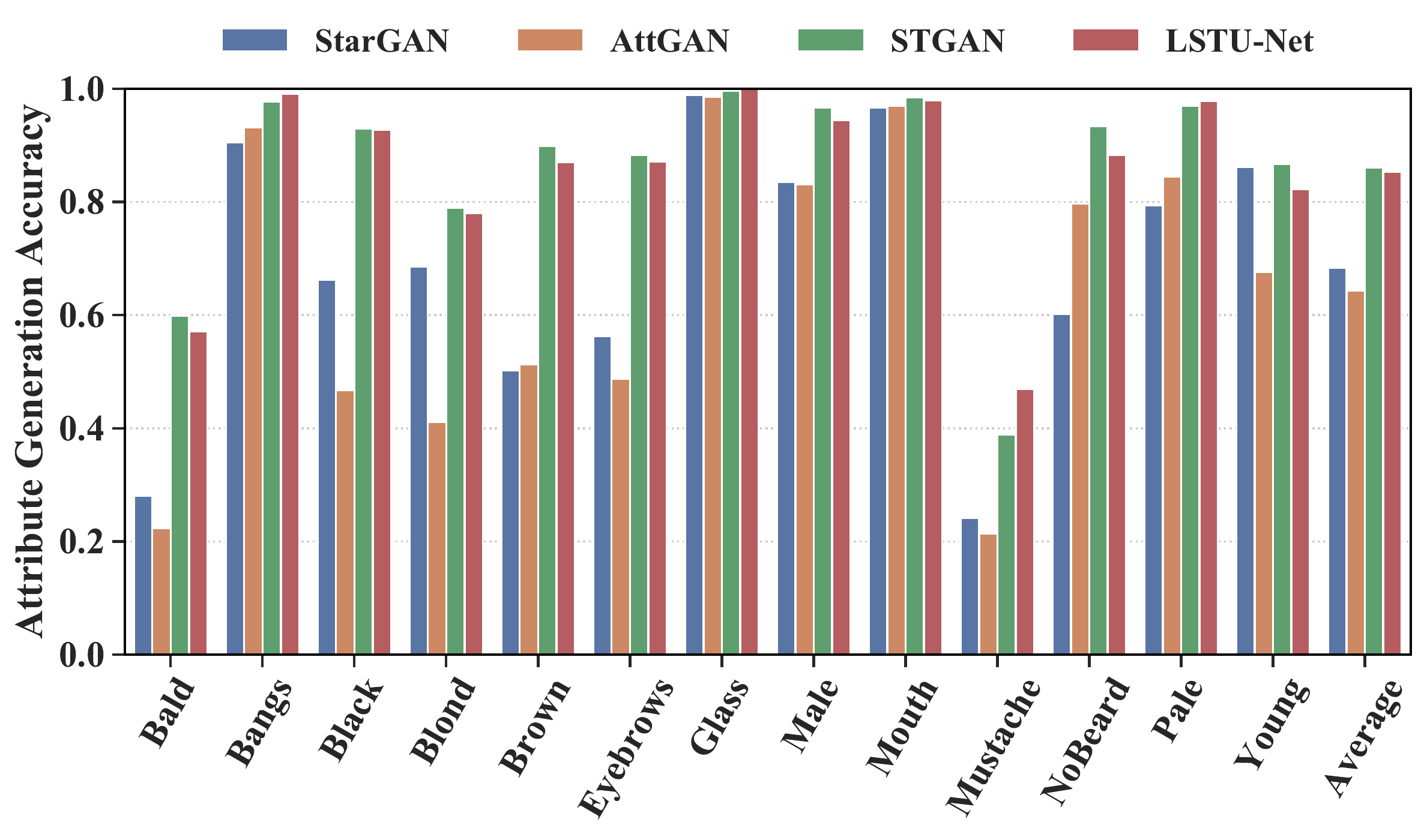}
\end{center}
  \caption{Our LSTU-Net against other approaches in attribute translation accuracy.}
\label{acc1_LSTUnet}
\end{figure}

\begin{table}[t]
\begin{center}
\footnotesize
\setlength{\tabcolsep}{3.5mm}
\label{parameters2}
\begin{tabular}{ccccccc}
\hline
{channel}  & \multicolumn{2}{c}{16\ in-channel} &  \multicolumn{2}{c}{32\ in-channel}& \multicolumn{2}{c}{64\ in-channel} \\ 
\cline{1-7}
{metric}& param & FLOPS & param & FLOPS & param & FLOPS \\
\hline 
{STU}  & 1.63M & 5.06G & 6.40M & 19.99G & 25.37M & 79.50G \\
\hline
{LSTU}  & 0.81M & 2.02G & 3.20M & 7.85G & 12.68M & 30.80G \\
\hline
\end{tabular}
\caption{The total number of parameters and FLOPS in the skip connections within a five-layer network with different input channels for the LSTU and STU. G denotes GFLOPS.}
\label{complexity}
\end{center}
\end{table}

{\bf Quantitative evaluation.}
The attribute editing accuracy and image fidelity comparisons are shown in Fig.~\ref{acc1_LSTUnet} and Table~\ref{diffframeworkpsnr}. For attribute accuracy, our LSTU-Net outperforms StarGAN, AttGAN and ties with STGAN. 
This circumstance remains the same for image fidelity manifested through PSNR and SSIM. These results show that LSTU has the same level of effect as STU, whether in attribute editing accuracy or image fidelity. 
The complexity and computational cost comparisons of different units are demonstrated in Table~\ref{complexity}. Compared to STU, our LSTU reduces parameter size by half and FLOPS by over 60 $\%$.
These comparisons prove the effectiveness and efficiency of implementing LSTU.

{\bf Qualitative analysis.} The visual effect comparison of $128\times 128$ images is given in Fig.~\ref{lrresults}. The editing result of our LSTU is better than StarGAN, AttnGAN, and achieves similar visual effect with STGAN. In almost all attributes, our model not only performs thorough and accurate translation, but also suffer the least from textural artifacts. In contrast, the results of StarGAN have distinct color difference due to the lack of skip connection. AttGAN's results contain fake artificial textures and half-changed features in attributes like bald and hair color. Such inconsistency is caused by the direct use of skip connection. STGAN is more successful than AttGAN and StarGAN. And its edited features and ours are equally effective.
These comparison results present the state-of-the-art level attribute editing ability and tremendous efficiency advantage of our LSTU. More results generated by the LSTU-Net are given in the suppl.

\section{Conclusion}
In this paper, we study the problem of high-resolution facial attribute editing in resource-constrained conditions. By proposing the patch-based local-global framework \emph{CooGAN} along with the multi-scale features fusion method LSTU, we attained the high-resolution facial attribute editing with constrained GPU memory. We use an up-to-down approach for patch processing to preserve global consistency and most importantly, lower the computational resource requirement. And the LSTU retains the attribute variety in skip connection with fewer parameters and lower computation cost. Experiments on facial attribute editing exhibit the superior performance of our framework in quality, attribute accuracy and efficiency to state-of-arts in the scope of facial attribute editing with constrained computational resources. Theoretically, our framework has no image size limitation owing to its patch processing method. We believe it has promising prospect not only in resource-constrained situations, but also in extremely high-resolution image processing tasks.

\section{Acknowledge}
This work was supported by National Science Foundation of China (61976137, U1611461, U19B2035) and STCSM(18DZ1112300).

\clearpage
%
%
\bibliographystyle{splncs04}
\bibliography{1426}
\end{document}